\documentclass[nohyperref]{article}

\usepackage{microtype}
\usepackage{graphicx}
\usepackage{subfigure}
\usepackage{booktabs} 
\usepackage{enumitem}

\usepackage{hyperref}

\usepackage[accepted]{icml2022}

\usepackage{amsmath}
\usepackage{amssymb}
\usepackage{mathtools}
\usepackage{amsthm}

\usepackage[normalem]{ulem}

\usepackage[capitalize,noabbrev]{cleveref}
\usepackage{enumitem,amssymb}
\newlist{todolist}{itemize}{2}
\setlist[todolist]{label=$\square$}
\usepackage{pifont}
\theoremstyle{plain}

\theoremstyle{definition}

\theoremstyle{remark}

\usepackage[textsize=tiny]{todonotes}
\newcommand{\xhdr}[1]{\vspace{1.7mm}\noindent{{\bf #1}}}

\icmltitlerunning{CADA-GAN: Context-Aware GAN with Data Augmentation}

\begin{document}

\twocolumn[
\icmltitle{CADA-GAN: Context-Aware GAN with Data Augmentation}



\icmlsetsymbol{equal}{*}

\begin{icmlauthorlist}
\icmlauthor{Sofie Daniëls}{equal,ETH}
\icmlauthor{Jiugeng Sun}{equal,ETH}
\icmlauthor{Jiaqing Xie}{equal,ETH}

\end{icmlauthorlist}

\icmlaffiliation{ETH}{Computer Science, ETH Zurich}

\icmlkeywords{Machine Learning}

\vskip 0.3in
]
 


\printAffiliationsAndNotice{\icmlEqualContribution}  

\section*{\centering Abstract}

Current child face generators are  restricted by the limited size of the available datasets. In addition, feature selection can prove to be a significant challenge, especially due to the large amount of features that need to be trained for. To manage these problems, we proposed CADA-GAN, a \textbf{C}ontext-\textbf{A}ware GAN that allows optimal feature extraction, with added robustness from additional \textbf{D}ata \textbf{A}ugmentation. CADA-GAN is adapted from the popular StyleGAN2-Ada model, with attention on augmentation and segmentation of the parent images. The model has  the lowest \textit{Mean Squared Error Loss} (MSEloss) on latent feature representations and the generated child image is robust compared with the one that generated from baseline models.

\begin{figure*}[t]
    \centering
    \includegraphics[width=\textwidth]{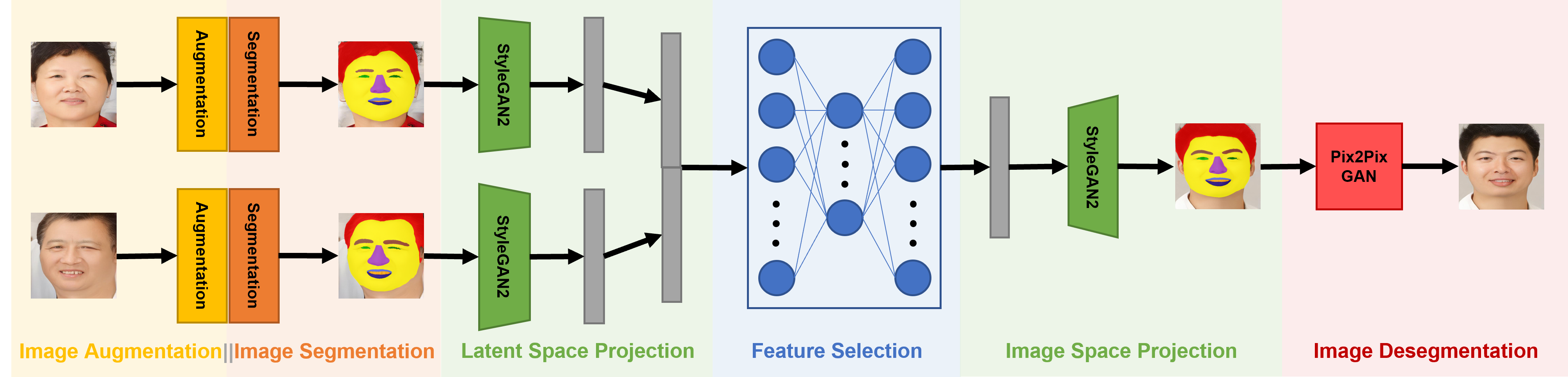}
    \caption{Pipeline of context-aware GAN with data augmentation.}
    \label{fig:01}
\end{figure*}

\section{Introduction}
\begin{figure*}[t]
\centering
\begin{minipage}[t]{0.4\linewidth}
\centering
\includegraphics[width=0.8\textwidth]{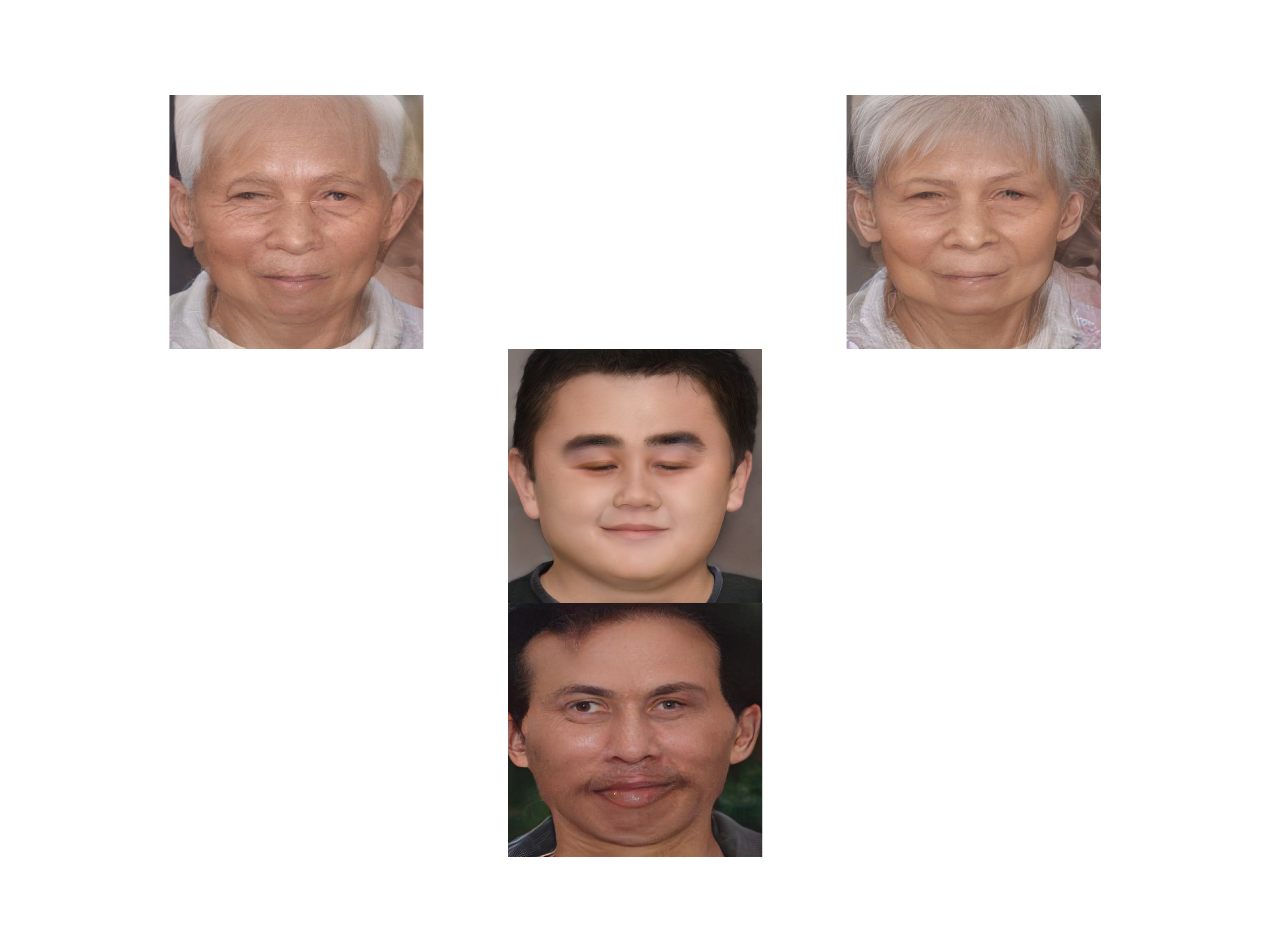}
\caption{Generated Child Images (Baseline)}
\label{2}
\end{minipage}
\begin{minipage}[t]{0.4\linewidth}
\centering
\includegraphics[width=0.8\textwidth]{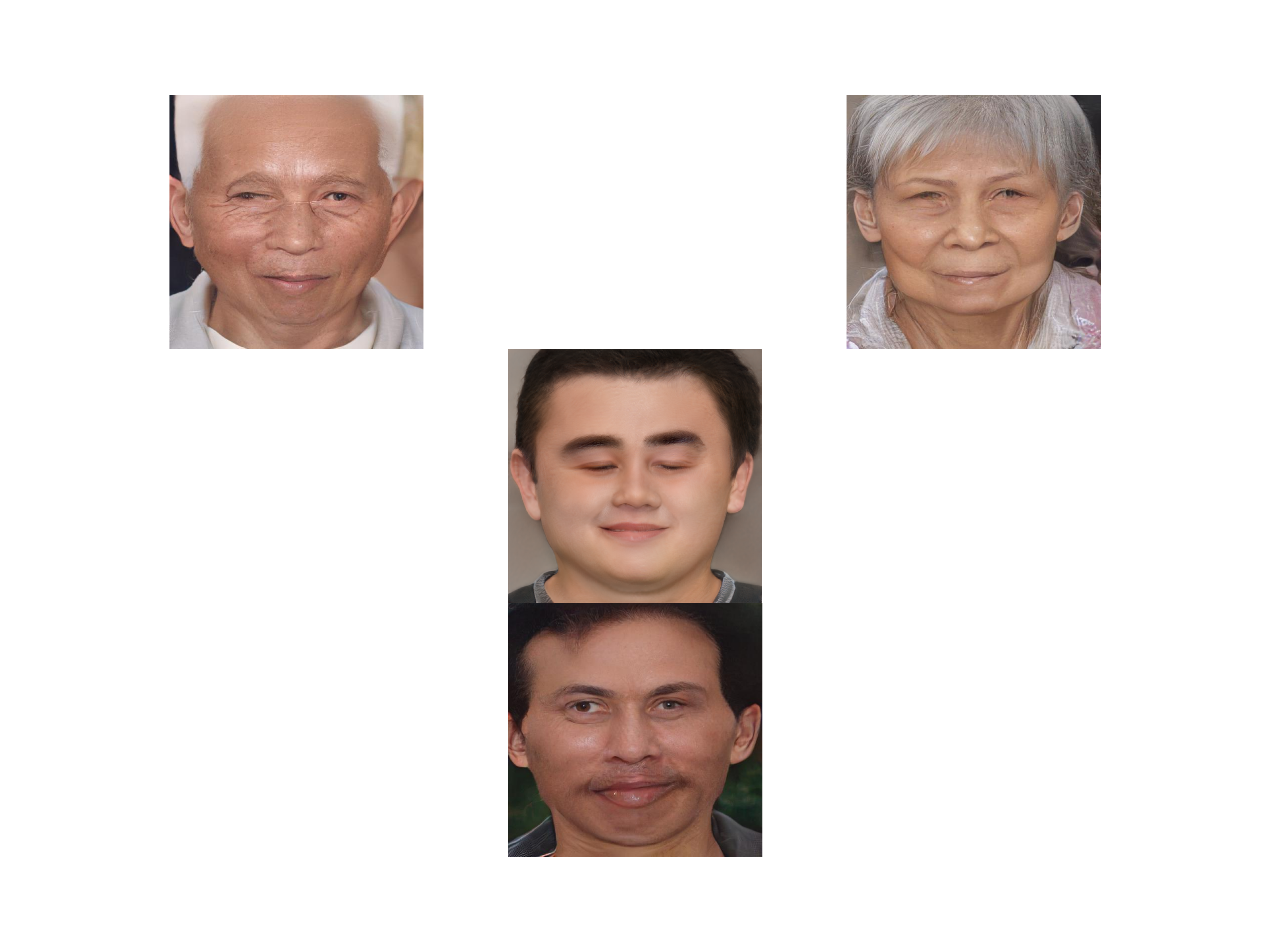}
    \caption{Generated Child Images (Mixup)}
\label{3}

\end{minipage}
\begin{minipage}[t]{0.4\linewidth}
\centering
\includegraphics[width=0.8\textwidth]{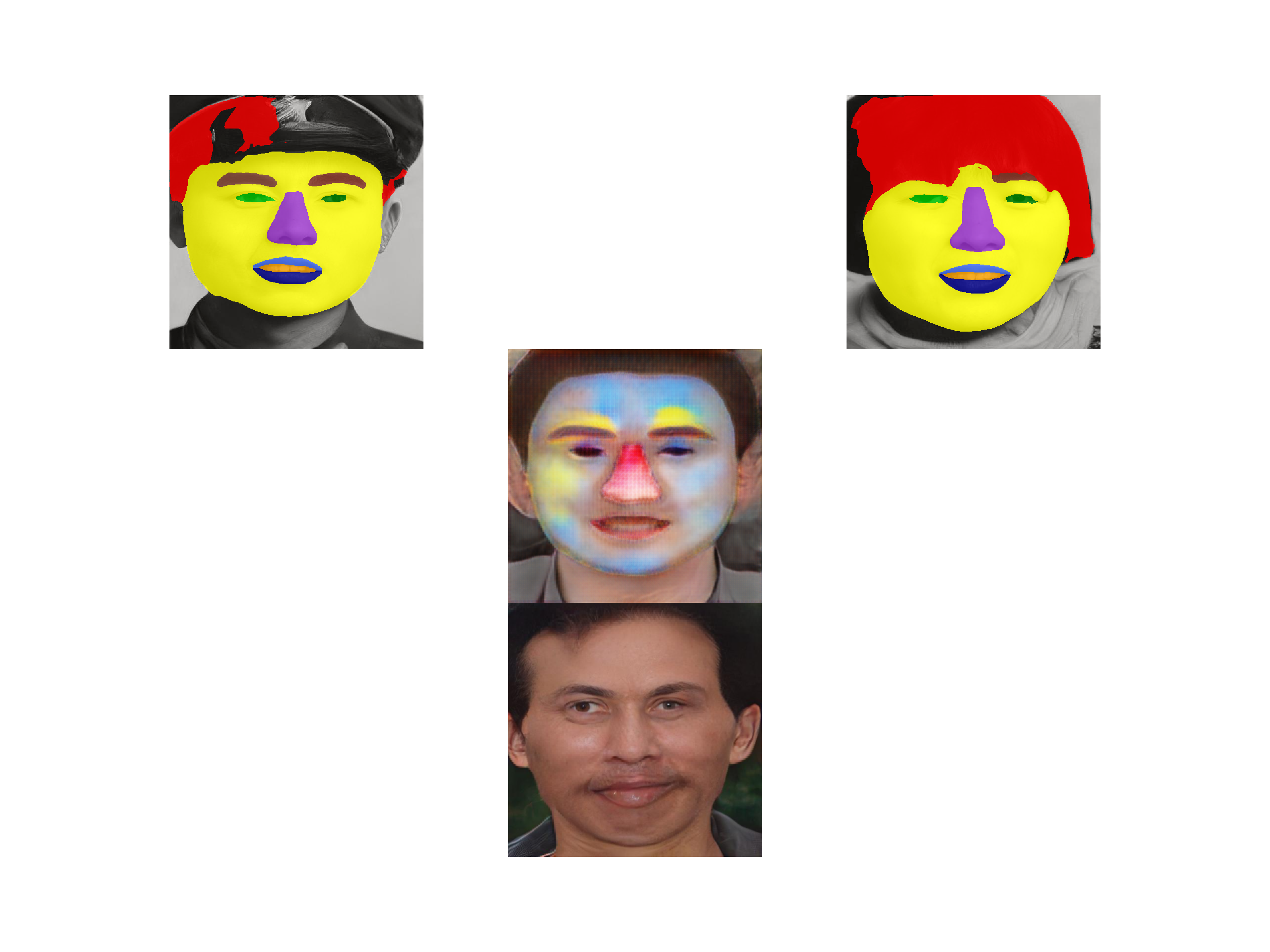}
    \caption{Generated Child Images (Segmentation)}
\label{3}

\end{minipage}
\end{figure*}

\label{introduction}
Generative adversarial networks (GANs) have become increasingly popular in recent years, in particular due to their ability to solve complex image-based problems, such as image generation \cite{bao2017cvae}, image-to-image translation \cite{liu2017unsupervised}, and face recognition \cite{tran2017disentangled}.
More specifically, GANs can be used to generate realistic faces in high-resolution \cite{wang2022gan, kammoun2022generative, wu2017survey}. One application of that is kinship face generation.

Prospective parents are typically highly interested in what their children will look like, and historians dream of automatically figuring out family trees from a limited number of portraits. Kinship face generation can greatly help with this.
In this paper, we specifically focus on the generation of faces of children using the images of their biological parents as input.

Some earlier papers rely on a one-to-one kinship relation \cite{SELFKIN2018}, but as better datasets are being published, more recent papers have allowed the use of multiple relatives as input \cite{Zhang20Face, Ghatas2020, gao2021dna, cui2021heredity, Lin2021StyleDNA}.

A major limitation of current works, however, is the limited dataset size. Obtaining father-mother-child triplets is difficult, and only few such datasets are available \cite{TSKinFace, FIW2016}. In addition, some datasets are highly pre-processed and do not allow generalization of existing algorithms to less homogeneous datasets.

To combat this, we propose two additional pre-processing steps. We first augment the images with various existing data augmentation techniques, in order to improve robustness of our model. In addition, we segment the images of the parents to allow the adversarial network to focus on the main features of the face.

The main contributions of this work are as follows:
\begin{enumerate}[noitemsep,topsep=0pt]
    \item We propose a model that improves robustness of child face generation by applying simple augmentations to the parent images.
    \item Using segmentation, our model additionally allows targeted highlighting of facial features for optimal child feature prediction.
    \item We employ transfer learning in order to facilitate downstream latent vector extraction of both parents and image reconstruction of the child.
\end{enumerate}

\section{Related Work}
\label{relatedwork}

\xhdr{Child Face Generation.}
Relatively few papers have been written on the topic of child face generation. Still, promising results were achieved using CDFS-GAN, GANKin, DNA-net, ChildGAN, and StyleDNA \cite{Zhang20Face, Ghatas2020, gao2021dna, cui2021heredity, Lin2021StyleDNA}.

CDFS-GAN is one of the first child face generators that uses both parents as input \cite{Zhang20Face}. By extracting main facial features, such as nose, mouth, and eyes separately, CDFS-GAN models feature-dependent genetic inheritance.

In contrast, GANKin directly extracts the features of the parents using FaceNET \cite{Schroff15Facenet} and uses a four-layer fully connected network to determine the child feature vector \cite{Ghatas2020}. The child face is generated from this feature vector using PGGAN \cite{karras_progressive_2018}.

DNA-net maps features to genes and back to incorporate genetic knowledge in the child face prediction \cite{gao2021dna}. The parent images are mapped to the feature space by training a conditional adversarial auto-encoder (CAAE) as proposed in \cite{Zhang17CAAE}. Features are selected at random, in order to model the real world more accurately. The final child face is generated using the decoder from the CAAE.

Similar to DNA-net, ChildGAN \cite{cui2021heredity} also relies on genetic knowledge to improve child face generation, but combines it with a semantic learning framework. First, the parent images are projected into the latent space using Image2StyleGAN \cite{Abdal2021}. After obtaining and finetuning the child vector with a macro- and microfusion step, the final child image is generated using StyleGAN \cite{karras2019style}.

\cite{Lin2021StyleDNA} instead use StyleGAN2 for latent space embedding. They compare an improved DNA-net based model, StyleDNA, with both a k-nearest neighbors and an eigenvector projection approach.

Most papers perform feature selection in the latent space to account for the imbalance between the limited dataset size and the number of learnable parameters. In this paper, we will therefore use StyleGAN2-Ada \cite{StyleGAN2ADA} for image-to-latent and latent-to-image space projection.

\xhdr{Image Augmentation.}
Data augmentation encompasses a range of techniques that increase the size of the dataset and improve robustness of training \cite{shorten_survey_2019}. Augmentation can be performed by deep neural networks or more fundamental image manipulation methods like CutOut \cite{devries_improved_2017} and MixUp \cite{hendrycks_augmix_2020}. One exciting strategy for augmentation is generative modelling, such as in \cite{radford_unsupervised_2016}, \cite{karras_progressive_2018}, and \cite{mirza_conditional_2014}. GANs are a method for extracting new information from a dataset, according to \cite{bowles_gan_2018}. In Deep Learning research, the idea of meta-learning mainly relates to the idea of neural network optimization via neural networks \cite{shorten_survey_2019}, which has been utilised in the Data Augmentation field as well \cite{perez_effectiveness_2017,lemley_smart_2017,cubuk_autoaugment_2019}. Though its creative uses are arguably where neural style transfer is best recognized, it is also a fantastic tool for data augmentation \cite{johnson_perceptual_2016,gatys_neural_2015}. The techniques we are utilising are MixUp \cite{mixup} and AugMix \cite{hendrycks_augmix_2020}.

\xhdr{Image Segmentation.}
Segmentation of facial features has many applications in the field of computer vision, such as landmark detection, head pose estimation, recognition of facial expressions, and face recognition \cite{Khan20Review}. \par
Conditional random fields (CRFs) are commonly used as the basis for human face segmentation, either implemented as is \cite{Warrell09, Khan17} or in the form of a convolutional network \cite{Liu15, Zhou17Parsing}. Convolutional networks can also be combined with recurrent networks, such as in \cite{Liu17} and \cite{Zhou17Segmentation}. Many other algorithms can be used for facial feature extraction, like random forests \cite{Khan15} and support vector machines (SVMs) \cite{Khan18}. Still, most of the available segmentation models require extensive and error-prone preprocessing steps, in the form of either facial landmark detection, image cropping, or image alignment. In this paper, we therefore rely on the pretrained model of \cite{Lin21}, who propose a scale, rotation, and transformation equivariant model with competitive accuracy.

\section{Proposed method: CADA-GAN}

Our proposed model consists of three main steps: 1. data augmentation, 2. segmentation, and 3. face generation. The full pipeline can be found in Figure \ref{fig:01}. 

\xhdr{Image Augmentation.}
In the first step, we augment the images of the parents using either MixUp or AugMix.
MixUp \cite{mixup} generates a weighted combination of image pairs from the training data, while AugMix \cite{hendrycks_augmix_2020} mixes augmented images through linear interpolation. All images are augmented with a probability \textbf{P}.

\begin{figure}[h]
\centering
\includegraphics[width=0.22\columnwidth]{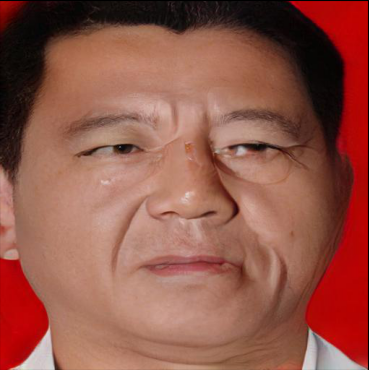}
\hskip -0.04in
\includegraphics[width=0.22\columnwidth]{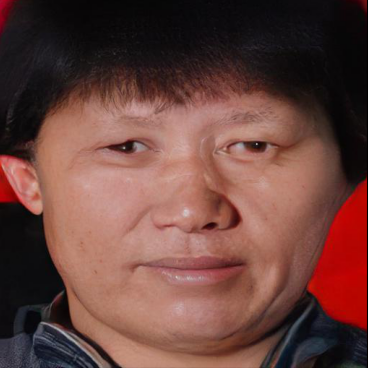}
\hskip 0.1in
\includegraphics[width=0.22\columnwidth]{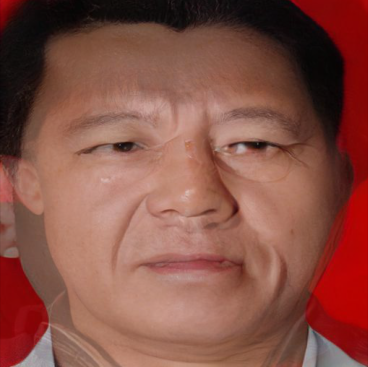}
\hskip -0.04in
\includegraphics[width=0.22\columnwidth]{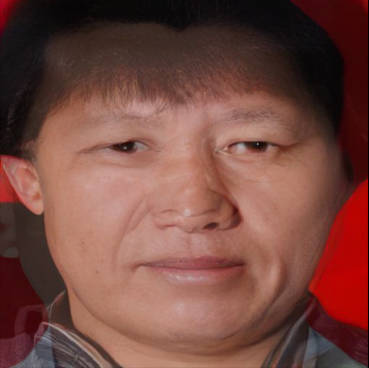}
\caption{Comparison of original images of a father and mother (left) with images augmented with MixUp (right).}
\label{fig:02}
\end{figure}

For MixUp, we follow the setup from \cite{mixup}: 
\begin{equation}
\left\{
             \begin{array}{lr}
             \texttt{img}_{f}' = \alpha * \texttt{img}_{f} + (1-\alpha) * \texttt{img}_{m} &  \\
             \texttt{img}_{m}' = \beta * \texttt{img}_{m} + (1-\beta) * \texttt{img}_{f} 
             &  
             \end{array}
\right.
\end{equation}
with $\texttt{img}_{f}$ and $\texttt{img}_{m}$ denoting the original images of the father and mother respectively. $\alpha$ and $\beta$ are drawn independently and at random from the interval [0,1]. Images of the children are not altered in any way.
The goal of MixUp is to approximate the feature space by combining as many instances of the input feature vector as possible, thus strengthening the generality of the generation. For an example, see Figure \ref{fig:02}.

AugMix chooses an appropriate image manipulation method at random and applies it to the pictures of the parents. The available techniques are shearing, translation, small-degree rotation, and horizontal flipping.
\begin{equation}
    \texttt{Img} = \mathcal{F}(\texttt{Img})
\end{equation}
where $\ \mathcal{F}\in\{shear, translate, rotate, flip\}$.

Translation and shearing can be applied with respect to either the x-axis or the y-axis. We omit the more drastic modifications from the original literature \cite{hendrycks_augmix_2020}, as we intend to retain most of the facial features. Like in MixUp, we only apply transformations to the parent images.

\xhdr{Image Segmentation.}
In the second preprocessing step, we apply segmentation to facilitate the downstream extraction of the main facial features of the parents.
For this, we use the pretrained model from \cite{Lin21}, which segments images of faces into 11 classes: background, hair, nose, left eye, right eye, left eyebrow, right eyebrow, upper lip, lower lip, inner mouth, and skin. (Figure \ref{fig:04} in Appendix.)

\xhdr{Child Face Generation}
After preprocessing, we convert the segmented faces of the parents into latent vectors $\mathbf{g}_{m_{i}}$ and $\mathbf{g}_{f_{i}}$ using \cite{StyleGAN2ADA}, where $\mathbf{g}$() performs augmentation and segmentation, and embeds vectors to latent space. ($m_{i}$, $f_{i}$) is a pair of parent images sampled from dataset $M \odot F$. We first finetune a pretrained model trained on the FFHQ dataset with images from the TSKinFace Dataset. Considering the slow convergence of tuning pretrained parameters on a new model, we apply transfer learning with a training session of 15 ticks.
The overall training loss for generating a child image in the latent space is:
\begin{equation}
    \texttt{rloss} = \sum_{(m_{i}, f_{i}, c_{i})
    \atop
    \in \{M, F, C\} } 
    \mathcal{L}(G(\texttt{aggr}(\mathbf{g}_{m_{i}}, \mathbf{g}_{f_{i}})), \mathbf{c}_{i}),
\end{equation},
where $G$ is the generator from the GAN and  $\mathcal{L}$ is the mean squared error (MSE) loss. Similar to GANkin \cite{Ghatas2020}, we then use a MLP with two fully connected layers (see Table \ref{tab:MLP}) as our aggregation function $\texttt{aggr}$ to predict the latent features of the child. The MSE loss is calculated between the original and predicted latent vectors of the child. We train with a batch size of 16 and use Adam as the optimization algorithm with a learning rate of 
0.00001.

\begin{table}[b]
\small
\centering
\resizebox{\columnwidth}{!}{%
\begin{tabular}{lll}
\hline
\textbf{Layer}           & \textbf{Input Size}          & \textbf{Output Size}        \\ \hline
\textbf{Fully Connected} & (None, 2x16x512) &(None, 512)     \\
\textbf{ReLU}            &                         &                        \\
\textbf{Dropout (p=0.25)}   &                         &                        \\
\textbf{Fully Connected} & (None, 512) & (None, 1x16x512) \\ \hline
\end{tabular}}
\caption[MLP Layers]{Layers of MLP used for feature extraction. \label{tab:MLP}}
\end{table}

\xhdr{De-Segmentation.}
To reverse the predicted segmentation of the child face into a real image at the end of the pipeline, we rely on a pix2pix GAN as proposed in \cite{Isola2017}. To account for the color variations in the StyleGAN2 output, we add a random value \textbf{x} to the hue of the original image and modify the saturation by \textbf{y}. We restrict the range of \textbf{x} and \textbf{y} to [-5, 5].
\begin{equation}
    \texttt{Img}_{h}' = \left( \texttt{Img}_{h} + \mathbf{x} \right) \;\mathrm{mod}\; 180
\end{equation}
\begin{equation}
    \texttt{Img}_{s}' = \min{ \left(255, \max{\left(0, \texttt{Img}_{s} + \mathbf{y} \right)}\right)}
\end{equation}

\xhdr{Datasets.}
We train and test the entire model with the TSKinFace Dataset \cite{TSKinFace}. TSKinFace is grouped into three family compositions, Father-Mother-Daughter (FM-D), Father-Mother-Son (FM-S), and Father-Mother-Son-Daughter (FM-SD), which allows straightforward extraction of the parent and corresponding child images. For computational reasons, we will only work with the FMS dataset.

The problem of poor reconstruction results from StyleGAN on low-resolution data inspires us to first transform our images to a high resolution image (512x512). We take advantage of SPARNET \cite{ChenSPARNet} to perform such transformations.

The pix2pix GAN is trained on images from the Flickr-Faces-HQ (FFHQ) datasets \cite{Lin21, karras2019style}. All images are resized to a resolution of 256x256 and segmented using the model from \cite{Lin21}.
Due to the vast size of the FFHQ-dataset, we only use 10 000 images to train the pix2pix GAN and are able to segment 9 872 correctly.

\xhdr{Code and Execution.}
All code was executed in the High Performance Computing Clusters of ETH Zürich on an NVIDIA GeForce GTX 2080 Ti GPU, where the full pipeline can be trained and tested in 18 hours on average. We also tested our code on an NVIDIA A100 GPU, where training of the baseline took only 15 hours.

\section{Evaluation}

\xhdr{Transfer learning}
15 ticks are performed according to our transfer learning model. At the very beginning, generated parents and child images are unlike to the original individual input since the pretrained model has learned nothing from the dataset. After 15 ticks, the models are retrained well so we are convinced to take it as the new pretrained model to be further fine-tuned quickly \ref{appendix:tl}.

\xhdr{Experiments} For evaluation, we compare four different scenarios.
Our baseline consists of the original GAN implementation with a four-layer network used for child feature prediction. In the first and second experiment, we apply image augmentation to the parent images in order to overcome the limited dataset size. We use either MixUp or AugMix for this. Both experiments did not show improvement when compared to the baseline, see Table \ref{tab:MSE}.
In the third experiment, we test the effects of segmentation and parse the parent faces with 11 labels. After feature prediction and image generation, the resulting child segmentation is converted back into a final realistic image with the pix2pix GAN. The results were significantly better than the baseline, as can be seen in Table \ref{tab:MSE}. The final experiment on the entire pipeline showed an averaged out result between segmentation and augmentation. Still, we believe the added robustness is worth the loss in accuracy.

\begin{table}[b]
\small
\centering
\resizebox{\columnwidth}{!}{%
\begin{tabular}{lll}
\hline
\textbf{Methods}           & \textbf{MSE (lat)}          & \textbf{MSE (images)}        \\ \hline
\textbf{StyleGAN2}  & 2.45 & \textbf{3372.53}   \\            
\textbf{StyleGAN2 + Mixup}       &  2.46 &  3485.57                       \\
\textbf{StyleGAN2 + AugMix}  & 2.47 & 3481.18                   \\
\textbf{StyleGAN2 + Segmentation}  & 2.06 & 3343.21                   \\
\textbf{CADAGAN (Ours)} & \textbf{2.17} & 5518.13 \\ \hline
\end{tabular}}
\caption[MLP Layers]{Validation MSE loss for both latent vectors and original images \label{tab:MSE}}. During the training, we do consider the similarity between latent vectors. Therefore the MSE loss would become smaller.
\end{table}

\xhdr{Evaluation Metric} Many papers measure the cosine similarity between the predicted latent vector of the child and the latent vectors of the parents, and compare it with the cosine similarity between the real child and parental latent vectors \cite{Ghatas2020, cui2021heredity, gao2021dna}.
The cosine similarity equation is given by:
\begin{equation}
    \texttt{cos}_\texttt{sim} = 1 - \frac{\textbf{u}\cdot \textbf{v}}{\lVert u \rVert \lVert v \rVert }
\end{equation},
as per the official SciPy documentation, with $u$ and $v$ two vectors or flattened matrices. If they are similar to each other element-wise, then the subtracted term in equation 6 is equal to one, meaning that the lefthand term $\texttt{cos}_\texttt{sim}$ is equal to zero.
Recent papers consider it a success if the predicted latent vector is more similar to the parents than the original one. However, since some feature selection mechanisms automatically predict latent features similar to those of the parents, this metric does not account for the true genetic variety in the population. Therefore, we opt to evaluate the cosine similarity between the predicted and real child image. Since the main goal of child face generation isn't to generate as many different potential siblings as possible, we find that our metric is less subjective and less biased towards the feature selection method.
The results can be seen in Table \ref{tab:cos}.

\begin{table}[htb]
\scriptsize
\centering
\resizebox{\columnwidth}{!}{%
\scalebox{0.7}{
\begin{tabular}{ll}
\hline
\textbf{Methods}           & \textbf{cosine similarity}                \\ \hline
\textbf{StyleGAN2}  & 2 \\            
\textbf{StyleGAN2 + Mixup}       &  2         \\
\textbf{StyleGAN2 + AugMix}  & 2                  \\
\textbf{CADAGAN (Ours)} & \textbf{0.15} \\ \hline
\end{tabular}}}
\caption[MLP Layers]{Cosine Similarity between generated images and original images. The lower is better. The maximum value is 2. \label{tab:cos}}
\end{table}

\section{Conclusion and Future work}



We proposed CADA-GAN, a novel GAN model to deal with child face synthesis considering face segmentation and augmentation from original data. We tried different augmentation methods to expand our datasets, which brought invariance effects when applied to convolutional networks and strengthened the model. Applying segmentations lowered the general reconstruction loss in comparison to the baseline model. In addition, latent vector extractions for StyleGAN2 with a transfer learning model greatly speeds up training.

There could be two limitations according to our observations on CADA-GAN. We only tested two picture alteration techniques for the augmentation part. We might try some other methods based on equivariant or invariant operations on images, such as SmartAugment \cite{lemley_smart_2017}, which can be integrated into the pipeline.
In addition, instead of highlighting facial features with a  color segmentation, blurring of the non-important features, such as background, hair, and skin, could potentially further improve feature extraction, while simultaneously reducing the color artifacts produced by the pix2pix GAN.

Although our model did not perform adequately, we still believe in the potential added robustness and awareness of our proposed architecture.


\bibliography{example_paper}
\bibliographystyle{icml2022}







\end{document}